%% file: main.tex
\pdfoutput=1

\documentclass[11pt]{article}

\usepackage[]{acl}

\usepackage{times}
\usepackage{latexsym}

\usepackage[T1]{fontenc}

\usepackage[utf8]{inputenc}

\usepackage{microtype}

\usepackage{inconsolata}

\usepackage{graphicx}

\usepackage[normalem]{ulem}
\input{def}
\usepackage{amsmath}
\usepackage{subfigure}
\usepackage{booktabs}

\usepackage{floatrow}

\title{Multi-step Problem Solving Through a Verifier: An Empirical Analysis on Model-induced Process Supervision}

\author{Zihan Wang\textsuperscript{1}\thanks{\ \ This work was done during the author’s internship at Google.} $\ \ $ Yunxuan Li\textsuperscript{2}\thanks{\ \ Correspondence: yunxuanli@google.com} $\ \ $ Yuexin Wu$^2$ $\ \ $ Liangchen Luo$^2$ \\ \textbf{Le Hou}$^2$ \ \
\textbf{Hongkun Yu}$^2$ $\ \ $ \textbf{Jingbo Shang}$^1$\\
  $^1$University of California, San Diego $\qquad$ $^2$Google
}

\begin{document}
\maketitle
\begin{abstract}
\input{0-contents/01-abstract}
\end{abstract}

\input{0-contents/contents}

\bibliography{cite}

\appendix
\input{0-contents/appendix}

\end{document}

%% file: def.tex
\usepackage{graphicx}
\usepackage{xspace}
\usepackage[inline]{enumitem}
\newcommand{\method}{\mbox{MiPS}\@\xspace}
\newcommand{\smallmodel}{\text{PaLM 2-S}\@\xspace}
\newcommand{\largemodel}{\text{PaLM 2-L}\@\xspace}
\newcommand{\processverifier}{\mbox{PSV}\@\xspace}
\newcommand{\outputverifier}{\mbox{OSV}\@\xspace}

\newcommand{\sumodd}{\texttt{sum\_odd}\xspace}
\newcommand{\meanodd}{\texttt{mean\_odd}\xspace}

\newcommand{\sumlogit}{\texttt{sum\_logit}\xspace}

\newcommand{\sumlogprob}{\texttt{sum\_logprob}\xspace}

\newcommand{\maxv}{\texttt{max}\xspace}
\newcommand{\minv}{\texttt{min}\xspace}

%% file: 0-contents/01-abstract.tex
Process supervision, using a trained verifier to evaluate the intermediate steps generated by a reasoner, has demonstrated significant improvements in multi-step problem solving.
In this paper, to avoid the expensive effort of human annotation on the verifier training data, we introduce Model-induced Process Supervision (\method), a novel method for automating data curation.
\method annotates an intermediate step by sampling completions of this solution through the reasoning model, and obtaining an accuracy defined as the proportion of correct completions. 
Inaccuracies of the reasoner would cause \method underestimating the accuracy of intermediate steps, therefore, we suggest and empirically show that verification focusing on high predicted scores of the verifier shall be preferred over that of low predicted scores, contrary to prior observations on human curated data.
Our approach significantly improves the performance of PaLM 2 on math and coding tasks (accuracy +0.67\% on GSM8K, +4.16\% on MATH, +0.92\% on MBPP compared with an output supervision trained verifier). 
Additionally, our study demonstrates that the verifier exhibits strong generalization ability across different reasoning models.

%% file: 0-contents/contents.tex
\input{0-contents/11-introduction}
\input{0-contents/20-related}
\input{0-contents/30-method}
\input{0-contents/40-analysis}
\input{0-contents/50-conclusion}

%% file: 0-contents/11-introduction.tex
\section{Introduction}

Multi-step problem solving (e.g., math problems and coding challenges) showcases the capabilities of machine intelligence. 
While researchers have shown that model- and data-upscaling still hold powerful for large language models (LLMs) on multi-step problem solving~\cite{gpt4, llama2, gemini, selfimprove, llemma,wizardmath, metamath}, even the state-of-the-art LLMs still produce easily observable mistakes.
Furthermore, standard fine-tuning directly does not yield consistent and significant improvements~\cite{wizardmath,metamath,finetune}.

\begin{figure}[t]
    \centering
    \includegraphics[width=0.99\linewidth]{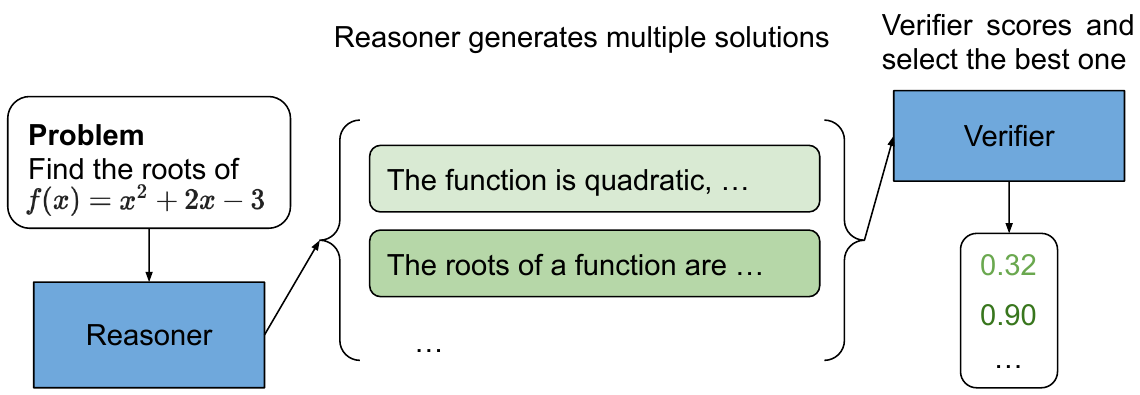}
    \caption{An illustration of the reasoner-verifier paradigm. The verifier predicts scores for the solutions generated by the reasoner, and selects the solution with the highest score. }
    \label{fig:reasoner_verifier}
\end{figure}

\begin{figure}[t]
    \centering
    \includegraphics[width=0.99\linewidth]{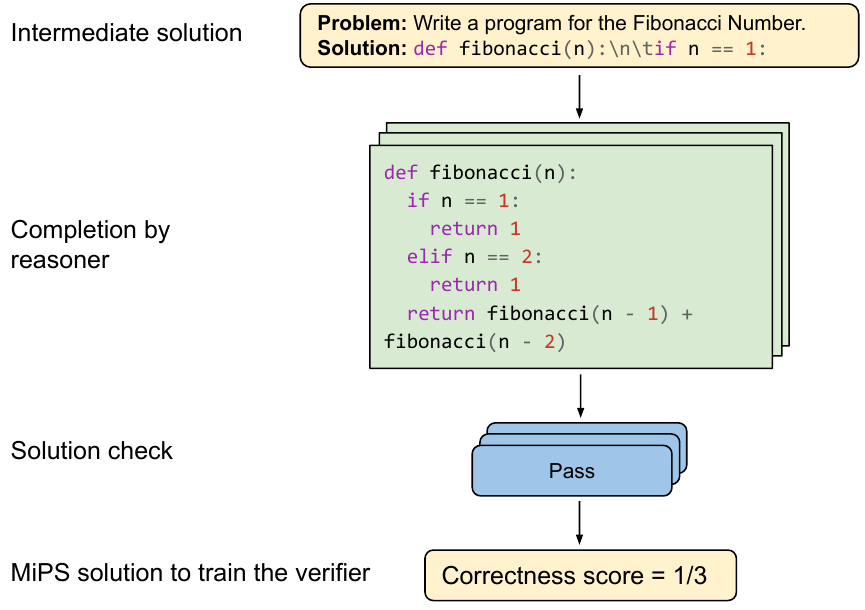}
    \caption{The Model-induced Process Supervision (\method) data construction method we introduce in this work. By completing an intermediate solution with a reasoner several times, we can obtain the percentage value of these completions being correct. These annotations are used to train a process supervised verifier.}
    \label{fig:mips}
\end{figure}

The reasoner-verifier paradigm (Fig.~\ref{fig:reasoner_verifier}) is as an inference-time technique where the goal is to pick one model-generated solution among many, since it is observed that there often are some correctly generated solutions. 
In particular, self-consistency~\cite{selfconsistency} is a special case of the verifier that picks the solutions that shares the majority answer with others (e.g., math tasks where the answer is a number). 
LLM-based verifiers are more general, as they could apply to arbitrary text solutions (e.g., code that implements a function)

Training a verifier in a supervised fashion has demonstrated strong performance in both coding and math language problems. 
\citet{gsm8k} showed that by simply gathering correct and incorrect solutions to train a binary classification model and using such model to pick the highest confidence solution generated by the reasoner during inference time, the accuracy can be improved significantly.
More recent studies suggest that verifying on intermediate steps could offer better guidance than training solely on the whole solutions~\cite{stepwise, processsupervision, refiner, stepbystep, tsllm, outcomeverification,  tinygsm, mathshepherd}. 
As such, verifiers trained (and applied) on intermediate steps are called process supervised verifiers (\processverifier), whereas those trained on whole solutions are called output supervised verifiers (\outputverifier).
In prior work, process supervision data is either obtained by an ad-hoc algorithm~\cite{stepwise, refiner}, or through expensive human annotations~\cite{processsupervision, stepbystep}, lacking an automatic and generic way of constructing of annotations of intermediate solutions.

Training verifier models require solution wise or step-wise labels, which is expensive to collect. There have been a series of work following an LLM-as-a-verifier approach where an off-the-shelf LLM is employed to judge the solutions through prompting~\cite{selfrefine, selfrefine2, selfcorrectsurvey}.
However, while such work may have seen improvements on language tasks, they haven't been very successful in math or coding problems~\cite{cannotselfcorrect, cannotcritique}.

In conclusion, to achieve optimal quality, training data is needed for building a strong verifier model. On the other hand, manually collecting solution verification labels is expensive and non-scalable. In this work, we propose to use Monte Carlo Sampling on the completions of the intermediate solutions to obtain step-wise training annotations (Fig.~\ref{fig:mips}).
Specifically, for each intermediate solution, we complete the solution with the reasoner several times through a sample decoding mechanism, and the percentage of the completed solutions being correct is referred to as the correctness of the solution. The correctness scores are used to train a \processverifier. 
Because of the nature of involving the reasoning model's completion on the intermediate solutions, we call the construction of this data Model-induced Process Supervision (\method). 
While such an idea is also explored in a concurrent work~\cite{mathshepherd}, we supplement with analysis of using \method constructed data. 
We find that because the reasoner model, which completes the solutions, is not perfect, the noises it introduces would affect the design choices of training and using the process supervised verifier:
\begin{itemize}[leftmargin=*,nosep]
\item We analyzed various ways to merge step-wise prediction scores to a single score value (we refer to this process as using an aggregation function) when using the verifier. Prior work used an aggregation function that focuses on low predicted scores and worked well for \processverifier trained on noise-free human annotated data~\cite{stepbystep}. For the noisy \method data, we suggest aggregation functions that focus on high predicted scores. 

\item We re-examine the usefulness of process supervision by isolating the trained \processverifier and studying the benefits of incorporating the predicted score from each intermediate step during verification. Our results reveal that (1) the verification scores from later intermediate steps are indeed useful even for a \processverifier trained on the noisy \method data, however, the earlier step scores could hurt the verification; and (2) only using the \processverifier predicted score of the last step, in similar fashion as \outputverifier, can sometimes be much better than \outputverifier itself, indicating process supervision data can regularize \outputverifier training. 

\item We show that verifiers trained on \method data generated by a reasoner can transfer to validate solutions by a different (and more competent) reasoner. This indicates that \method would not produce verifiers that are overly biased towards mistakes of the reasoner that generated the data. 
\end{itemize}

Following in this paper, we will provide a more complete review of related works, a precise definition of our method, and empirical results of the method and analysis on two math problem datasets and one coding dataset. The contributions of the paper are mainly (1) we propose \method to construct process supervision data automatically for training process supervision verifiers; (2) we extend the evaluation of problem solving verifiers to coding problems; (3) we provide empirical analysis on design choices and properties of the trained verifier from \method data. 

%% file: 0-contents/20-related.tex
\section{Related Works}

The advances of problem solving of LLMs can be broadly characterized into two regimes, first by \textit{training} a better reasoning model and the second by \textit{validating} the solution from the reasoning model at inference time.

\noindent\textbf{Pre-training/Fine-tuning Better Reasoners.}
Standard training recipes also transfer to training better reasoners for problem solving. During pre-training, larger model sizes and training compute yields an LLM that is more competent in multiple aspects~\citep[\textit{inter alia}]{gpt4, llama2, palm2}. 
Within fine-tuning, it is also observed that transfer learning~\cite{llemma} from a pile of generic math datasets, training on an augmented dataset of failure examples or diverse statements~\cite{selfimprove, wizardmath, metamath, finetune, dspy} leads to improvements.
Despite these approaches, it is apparent that (1) the state-of-the-art LLM still can fail at simple mistakes during multi-step problem solving and (2) the improvement of a simple verification method by majority voting (self-consistency~\cite{selfconsistency}) is still significant upon fine-tuning.
Therefore, the exploration of verifiers to validate and pick the solutions is necessary.


\noindent\textbf{Validating Through LLM-as-a-verifier.}
There have been numerous attempts on using the LLM reasoner itself to correct and validate its generated solutions. 
\citet{selfrefine, selfrefine2} and many methods surveyed in \citet{selfcorrectsurvey} broadly follow the strategy where the LLM validates and provides feedback to the generated solutions through prompts. 
\citet{cannotselfcorrect} and \citet{cannotcritique} revisited these methods and found that LLMs are not good verifiers for equally competent solutions, as such methods improves marginally on math word problems.

\noindent\textbf{Validating Through Trained Verifiers.}
In contrast, verifiers trained on a human labelled dataset does show significant improvements~\cite{gsm8k, processsupervision, stepbystep}. 
Importantly, \citet{stepbystep} showed that on a challenging competition-level mathematics problem set~\cite{math}, verifiers trained on annotated intermediate solutions (\processverifier) surpasses verifiers trained on final solutions by a large margin, and both substantially better than self-consistency. 
Other analysis also emphasized on the importance of step-wise feedback: \citet{processsupervision} showed that \processverifier selects solutions that are more accurate in their reasoning and \citet{tot, tsllm, mctsdecoding}, \textit{inter alia}, showed that during decoding, LLMs can be guided towards better solutions step-by-step.
We believe that improving the training of a \processverifier, and especially, identifying a scalable solution to generate the process supervision data, is of imminent importance.
Therefore, in this work, we identify an automatic and generic solution to generate process supervision data (\method), and conducted detailed analysis centered on the noises of this automatic process. 

\noindent\textbf{Math-Shepherd.}
Coincidentally, such an automatic process supervision data curation method was studied concurrently and independently by \cite{mathshepherd}. We share a generally similar methodology with their work, with a few minor design differences we highlight in later sections. The empirical results of \method are similar on the two datasets we share (GSM8K and MATH) despite using different, but about competent, LLMs. Their work extended training the verifier by applying it to fine-tune the reasoner through reinforcement learning, while our work included an additional coding dataset (MBPP) and provided analysis on the design choices of using the verifier, addressing the data noises. We believe these two works complement each other. 




%% file: 0-contents/30-method.tex
\section{Model-induced Process Supervision}

We consider the reasoner-verifier framework where we start with a fairly competent reasoner on a task, generate verifier training data on a given set of problems with the reasoner, and train a verifier on the data to validate some new generated solutions by a reasoner. 
We first discuss Model-induced Process Supervision, our data curation method that automatically creates process supervision data. 
Then, we discuss the details about the verification process.




\subsection{Obtaining MiPS data}
\method constructs process supervision data through Monte Carlo sampling.
First, we employ a reasoner model $r_g$ to generate a fix number of $n_g$ solutions for each problem, using temperature based decoding with a temperature of $t_g$. 
Then, for each solution, we decompose them into individual steps (we treat each line in a solution as an individual step). 
After that, for each intermediate solution containing a prefix list of steps, we employ a reasoner model $r_{mc}$ to generate again $n_{mc}$ solutions, with a temperature of $t_{mc}$, completing the intermediate solution.
For each completed intermediate solution, we calculate the percentage (out of $n_{mc}$) of them being correct, and these correctness values comprises the \method data. 
In all experiments in this paper, we consider $r_g = r_{mc}$, namely, the reasoner model that is used to estimate the intermediate solution's correctness is the same model that generates the solution data. This is particularly the most challenging case for \method, otherwise, using a more capable reasoner for the completion can enjoy a reduction of noise in \method data. 

\subsection{Training an Output Supervised Verifier}
To understand how well the process supervised verifier (\processverifier) trained from \method is, it is necessary to consider the vanilla output supervised verifier (\outputverifier), which uses the same amount of human labeling resources. 
The training data for \outputverifier are the generations from the reasoner $r_g$ with the same temperature value $t_g$. 
The verifier itself is a standard language model that supports a binary classification on the final token of the input. 
The verifier is trained on the cross entropy loss of the prediction.
This is also known as the solution-level verifier in \citet{gsm8k}.

\subsection{Training a Process Supervised Verifier}
The differences of training \processverifier and \outputverifier are:
\begin{itemize}[leftmargin=*,nosep]
    \item To enable predicting a score at each step in the solution, we mark the last token of each step (e.g., if each step is represented as a single line, the last token will be the new line token), and optimize step-wise predictions at each step at the same time. During inference, we would also obtain a score for each step in a solution.
    \item While for the output supervision data, or human labelled process supervision data, the score is either 0 or 1, for \method data, the correctness scores are percentage values. 
    The training objective considered in this work is to learn the exact percentage values $c_i$ for the $i$th step in the solution directly. 
    However, we note that it is possible to consider a different learning objective. For example, \citet{mathshepherd} considered learning a binarized score:
    \[   
         \Tilde{c_i} = 
         \begin{cases}
           \text{1,} &\quad\text{if }c_i > 0.0 \\
           \text{0.} &\quad\text{otherwise.} \\ 
         \end{cases}
    \]
    In later analysis, we compare these two objectives.
\end{itemize}

\subsection{Aggregating Step-wise Predictions}

\begin{table}[t]
    \small
    \centering
    \caption{We show the statistics of the datasets we use in this paper. The average number of steps is depicted with a granularity of 0.5, using \smallmodel for GSM8K and MBPP, and \largemodel for MATH. We note that these are not the most standard data splits, for reasons explained in Sec~\ref{sec:dataset}. }
    \label{tab:dataset}
    \resizebox{\linewidth}{!}{
    \input{1-tables/dataset}
    }
\end{table}

The trained verifier is used to score the solutions generated by the reasoner.
For \outputverifier, the verifier prediction can be directly used as the score for the solution. 
For \processverifier, the verifier predictions are a list of predicted probabilities $p_1, p_2, \ldots, p_n$, one for each step in the solution.
Aggregating the predictions into a final score is necessary.
\citet{stepbystep} considered two aggregation functions:
\begin{equation*}
\begin{split}
    & \minv = \min \{p_1, p_2, \ldots, p_n\}, \\
    & \sumlogprob = \sum_{i=1}^n \log p_i = \log \prod_{i=1}^n p_i,
\end{split}
\end{equation*}
They claimed that both are equivalently good aggregation functions. In later analysis, we show that for \method data, these two functions are underperforming for the trained verifier, while,
\begin{equation*}
\begin{split}
    & \maxv = \max \{p_1, p_2, \ldots, p_n\}, \\
    & \sumlogit = \sum_{i=1}^n \log \frac{p_i}{1 - p_i}, \\
    & \meanodd = \frac{\sum_{i=1}^n \frac{p_i}{1 - p_i}}{n},
\end{split}
\end{equation*}
are much better. We provide an analysis with a much larger set of aggregation functions and suggest that \method data prefers aggregation functions that focus on high prediction scores rather than lower ones. 

%% file: 1-tables/dataset.tex
\begin{tabular}{l|c|c|c}
\hline
Dataset & GSM8K & MATH & MBPP \\
\hline
Domain                       & math & math & coding \\
Fine-tuning \# Data       & 2000 & 4000 & 0   \\
Verification Training \# Data & 5000 & 8000 & 384 \\
Testing \# Data               & 1319 & 500  & 500 \\
Average Steps                 & 4.5  & 11.0 & 7.0 \\
\hline
\end{tabular}

%% file: 0-contents/40-analysis.tex
\section{Analysis}

\begin{figure*}[ht]
    \centering

    \subfigure[GSM8K, \smallmodel]{
        \includegraphics[width=.22\linewidth]{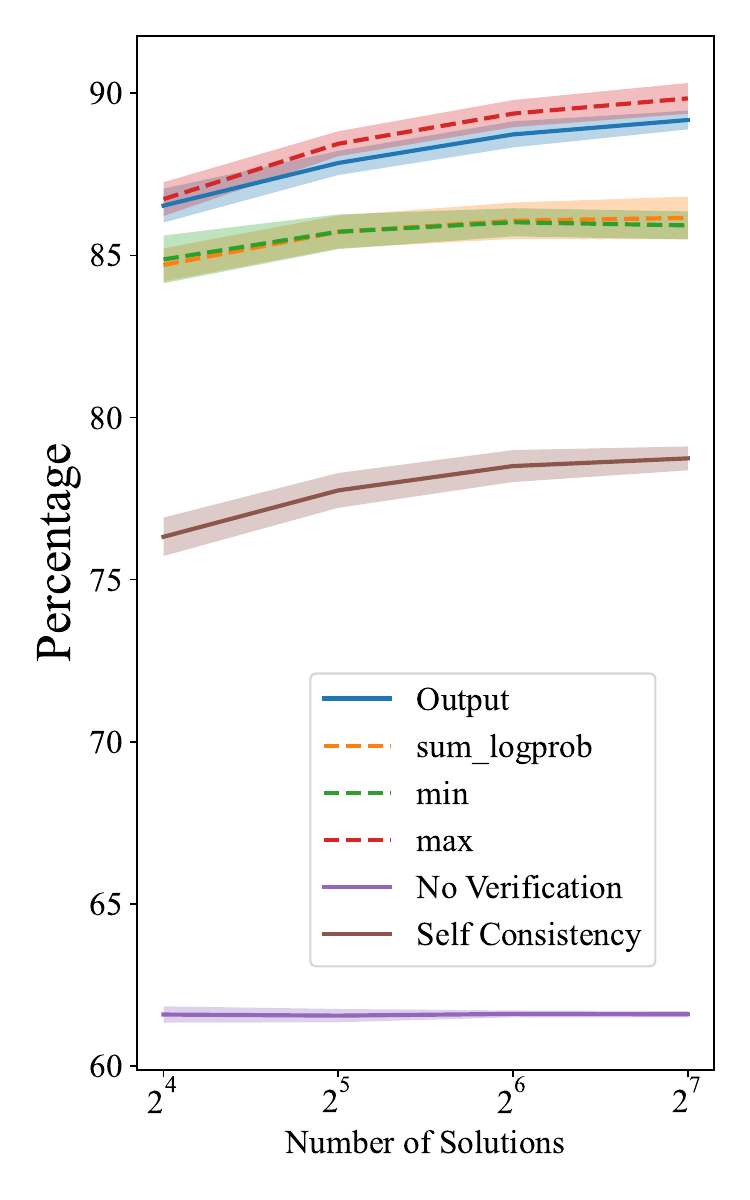}  
    }
    \subfigure[GSM8K, \largemodel]{
        \includegraphics[width=.22\linewidth]{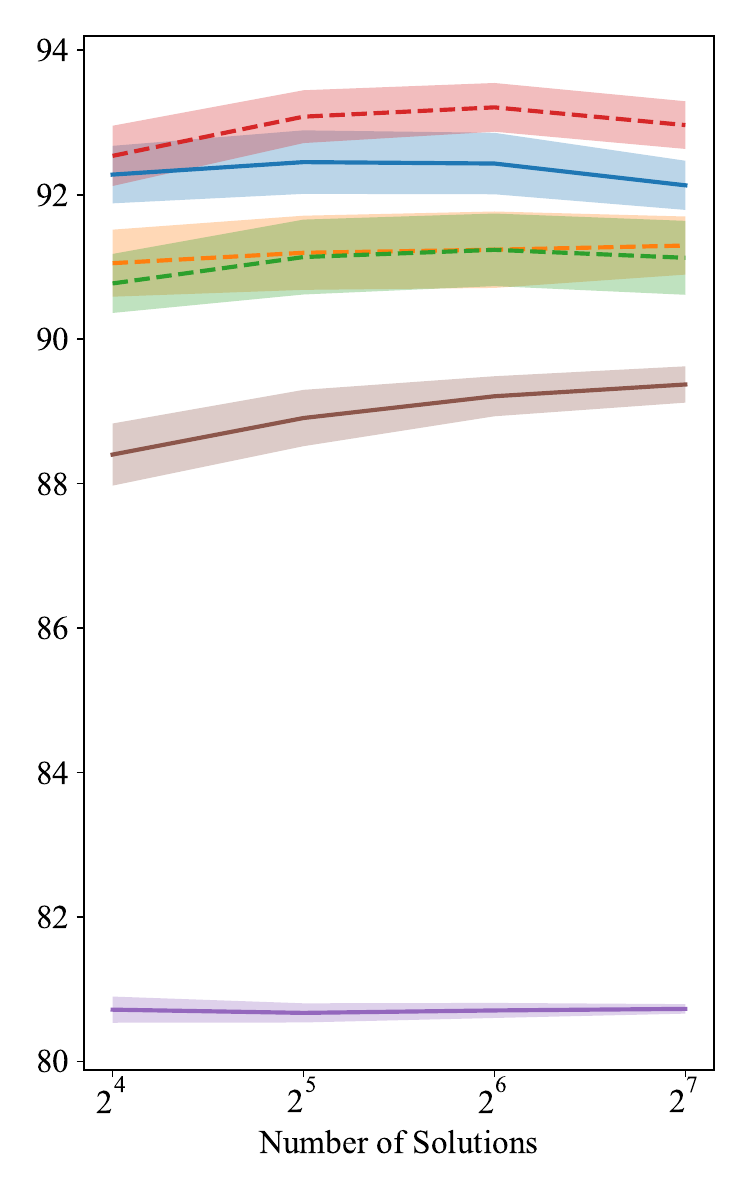}  
    }
    \subfigure[MATH, \largemodel*]{
        \includegraphics[width=.22\linewidth]{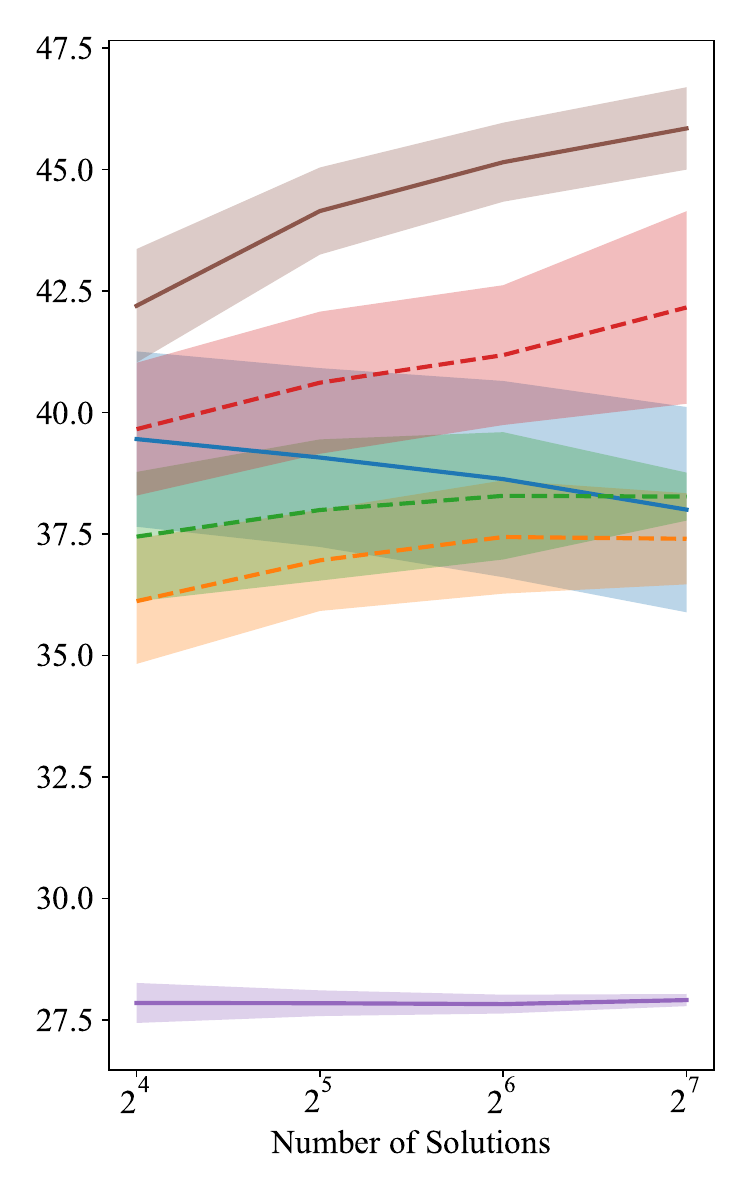}  
    }
    \subfigure[MBPP base model]{
        \includegraphics[width=.22\linewidth]{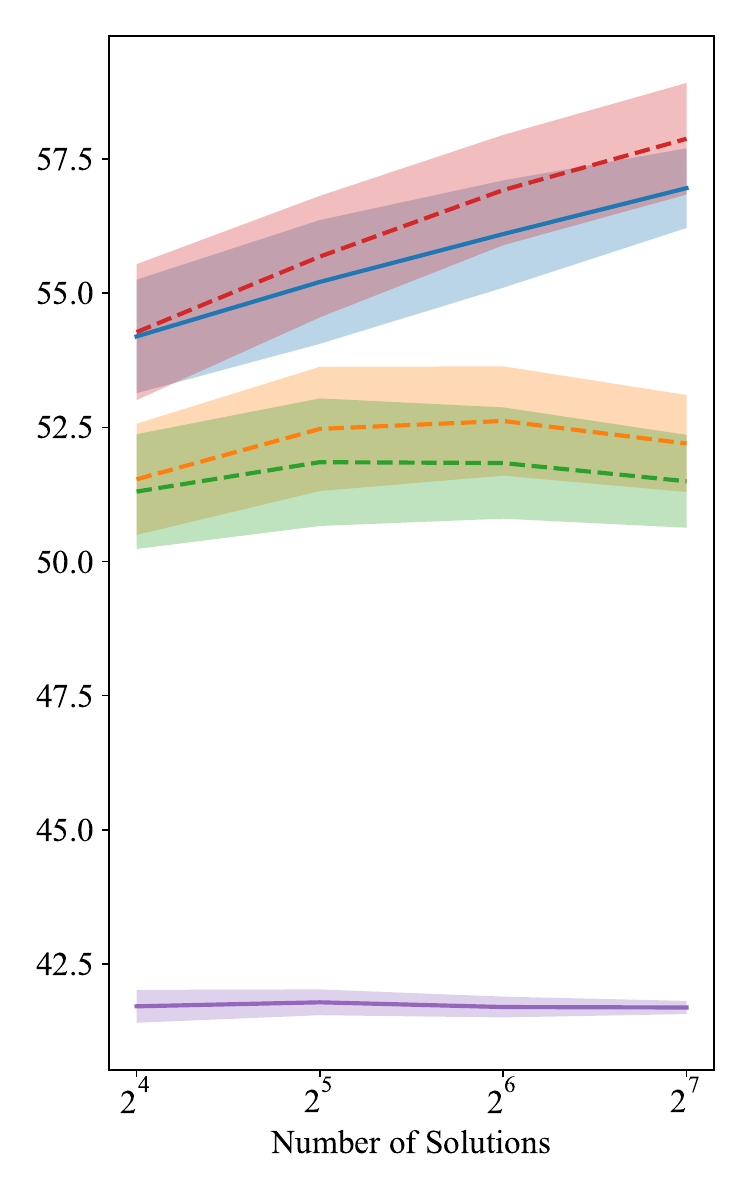}  
    }

    \caption{We apply the trained output- and process-supervised verifiers on various combinations of model and datasets. Self-consistency scores are given as a reference, however, it would be not applicable to general multi-step reasoning tasks (e.g., figure (d), coding). We use the default training objective that directly learns the estimated accuracies and the \maxv aggregation function for the process verifier. In the x-axis, we vary the number of generated solutions to apply the verifier on, and in the y-axis we plot the performance (accuracy \%). The standard deviation is also given. As a reference, we note that the purple line, representing the average performance of the generated solutions of the reasoner without any verification, matches the expectation to be an (almost) flat horizontal line with decreasing standard deviation. *While the reasoner that generates \method data and the reasoner that the verifier validates on is \largemodel, the verifier is trained from a \smallmodel. }
    \label{fig:main}
\end{figure*}

\subsection{Models}
In our experiments, we consider two LLMs, \smallmodel and \largemodel ~\cite{palm2} to conduct our experiments on. 
We intend to understand the capability of \method data and analyze design choices of the verifier when trained on it. 
A concurrent work~\cite{mathshepherd} conducted a similar experiment on a different set of LLMs, namely LLama2, LLemma, Mixtral, and Deepseek~\cite{llama2, llemma, mistral, deepseek}. 
Detailed experimental settings and hyperparameters can be found in Appendix~\ref{app:settings}.

\subsection{Datasets}
\label{sec:dataset}

We use two math datasets and one coding dataset for evaluations in this paper.
\begin{itemize}[leftmargin=*,nosep]
    \item \textbf{GSM8K}~\cite{gsm8k} is a dataset of grade school math problems. 
    \item \textbf{MATH}~\cite{math} is also a math word problems dataset. It consists of math problems of high school math competitions.
    \item \textbf{MBPP} is an entry-level Python programming dataset. The questions are coding challenges along with a test case that defines the function format and the solutions are Python code that is expected to solve several hidden test cases. 
\end{itemize}
Table~\ref{tab:dataset} contains detailed statistics about the datasets, and Appendix~\ref{app:datasets} contains more information on how we split these datasets into training and evaluation.

\subsection{Directly Applying MiPS}

\begin{figure*}[t]
    \centering
    \subfigure[GSM8K, \smallmodel]{
        \includegraphics[width=.3\linewidth,height=0.22\linewidth]{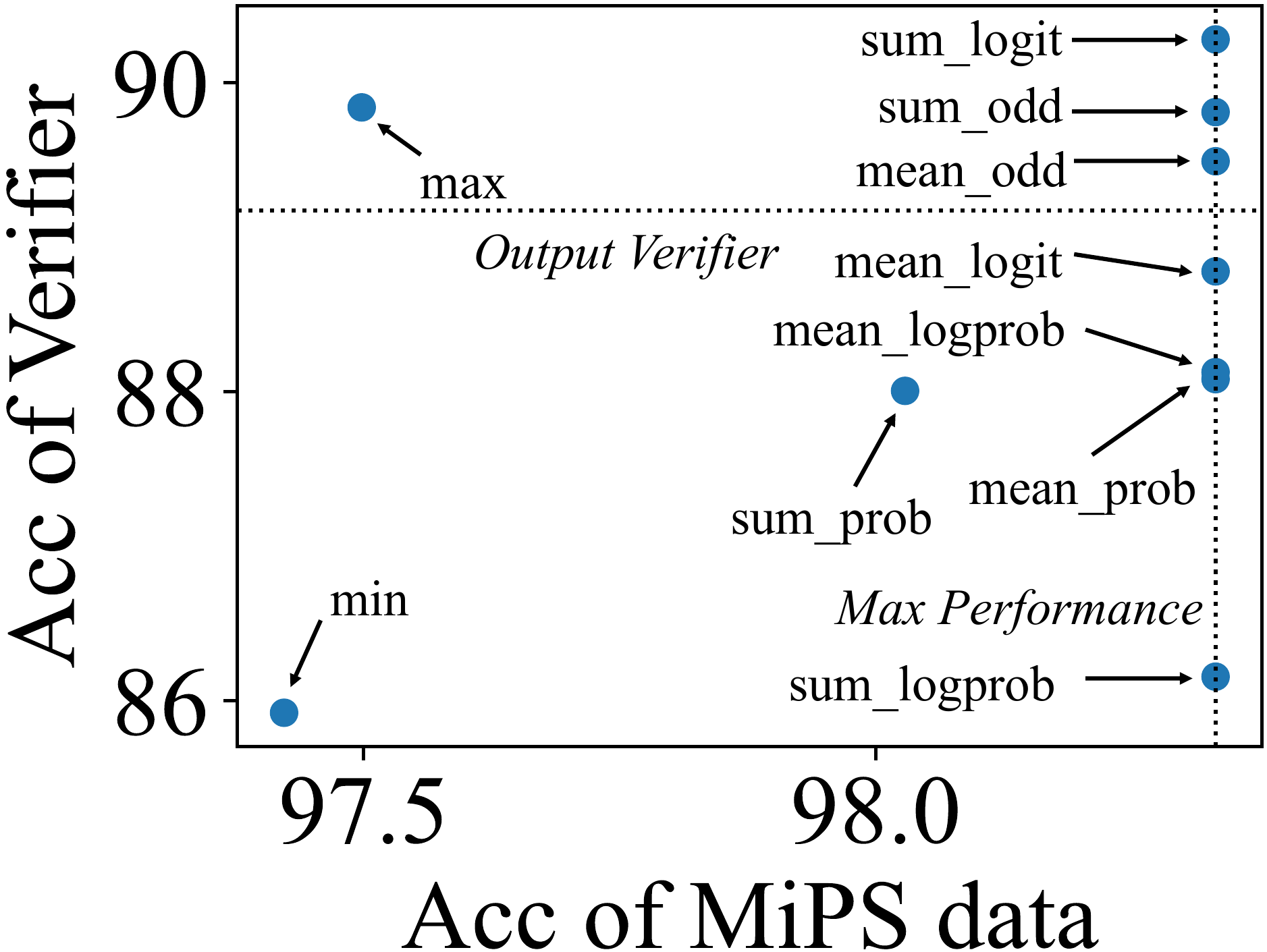}  
    }
    \subfigure[MATH, \largemodel]{
        \includegraphics[width=.3\linewidth,height=0.22\linewidth]{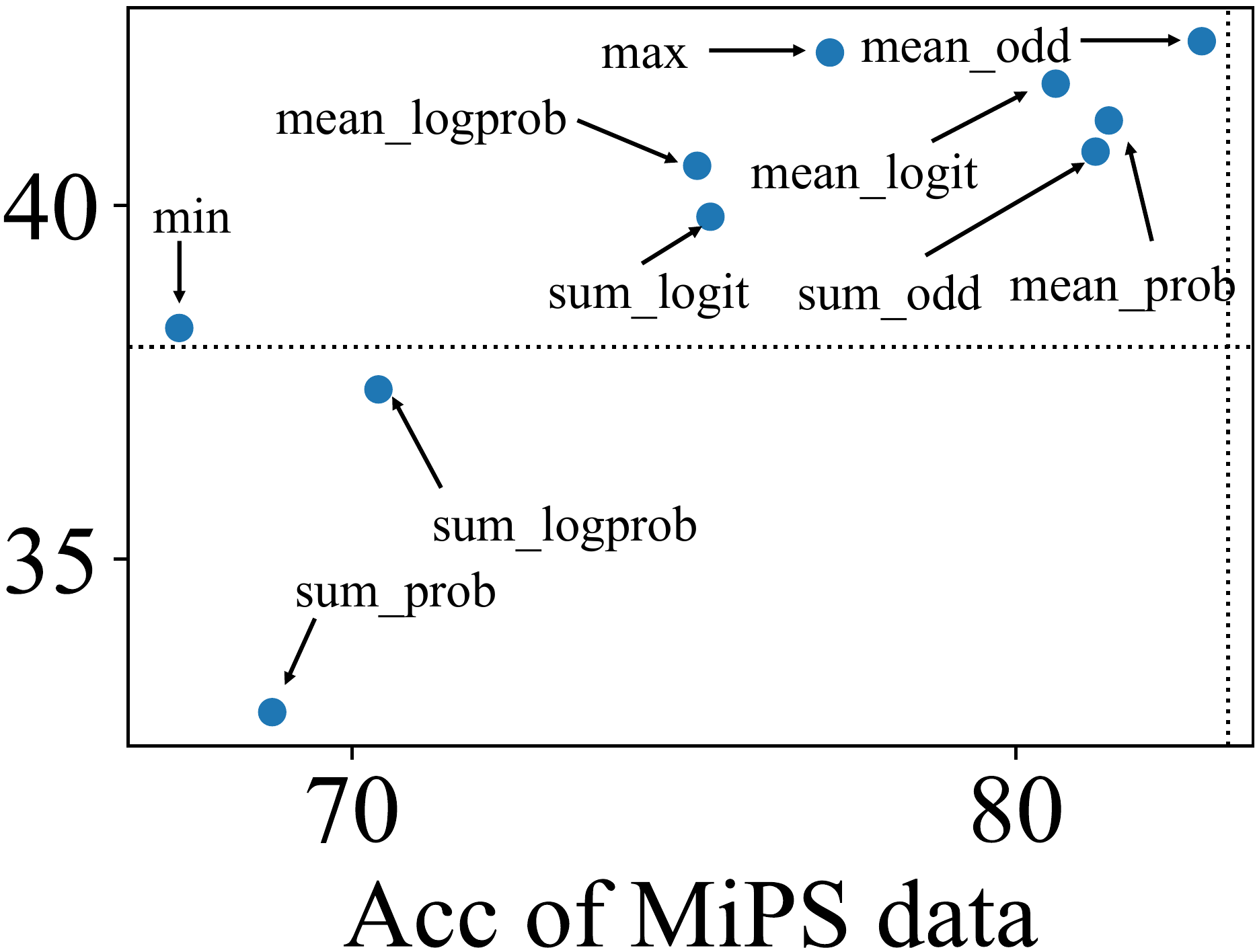}  
    }
    \subfigure[MBPP, \smallmodel]{
        \includegraphics[width=.3\linewidth,height=0.22\linewidth]{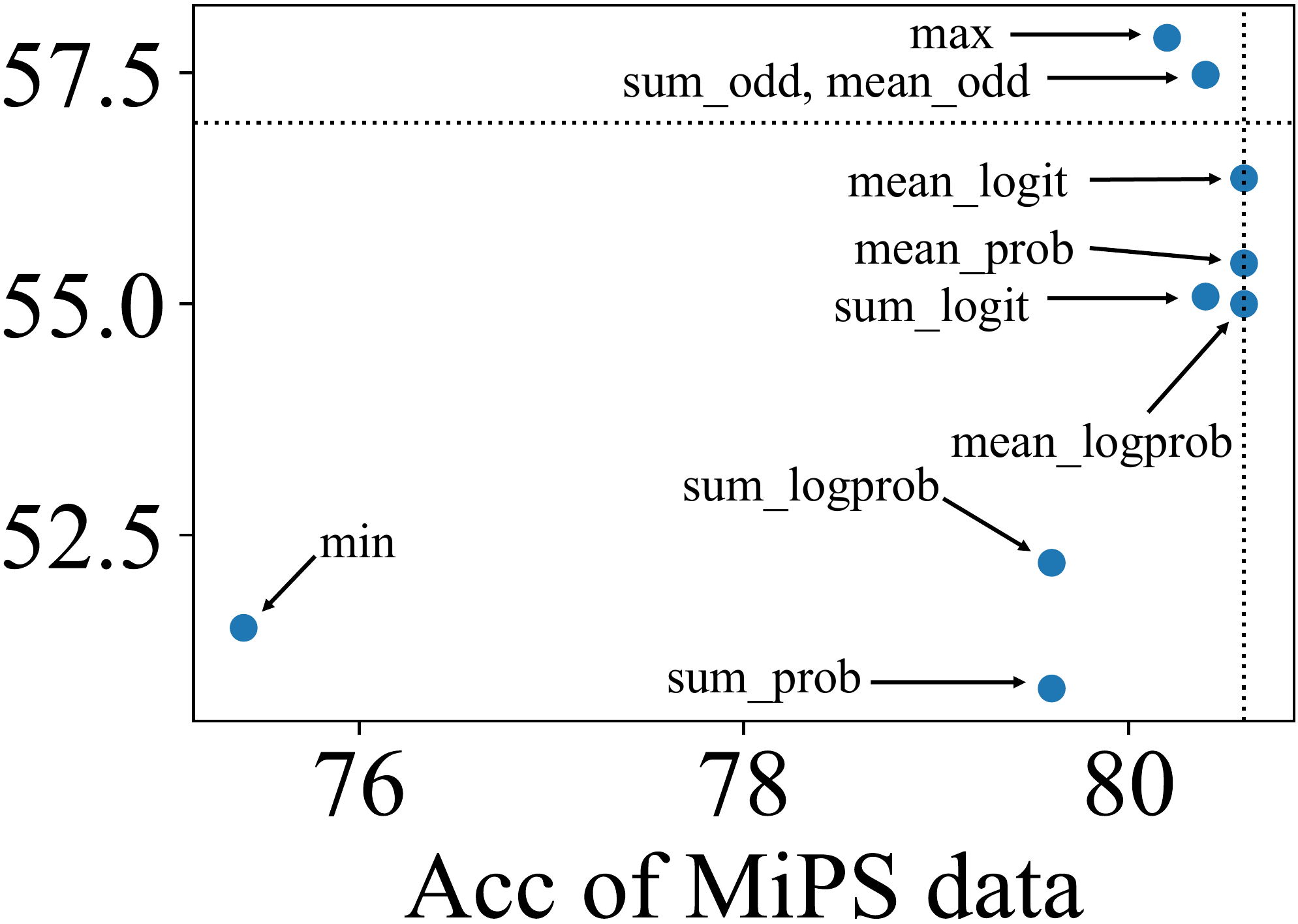}  
    }

    \caption{For each aggregation function, we show its accuracy on the \method data generated and the accuracy of using it with the \processverifier trained. We additionally plot two lines for easier understanding of the figure, a horizontal line corresponding to the performance of \outputverifier and a vertical line corresponding to the maximum accuracy achievable by the reasoner on the dataset (some problems are not solvable by the reasoner among the all solutions we generate). 
    }
    \label{fig:aggr}
\end{figure*}
\begin{figure*}[h!]
    \centering

    \subfigure[GSM8K, \smallmodel,\sumlogit]{
        \includegraphics[width=.3\linewidth]{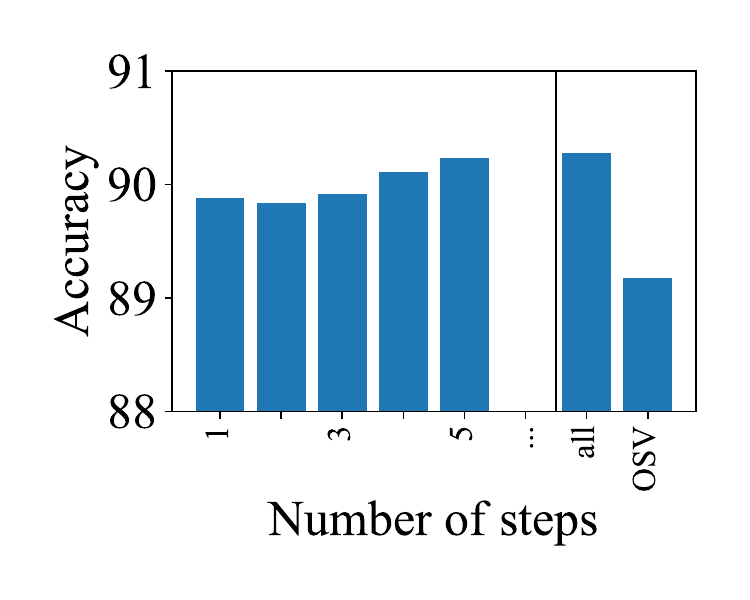}  
    }
    \subfigure[MATH, \largemodel, \sumlogit]{
        \includegraphics[width=.3\linewidth]{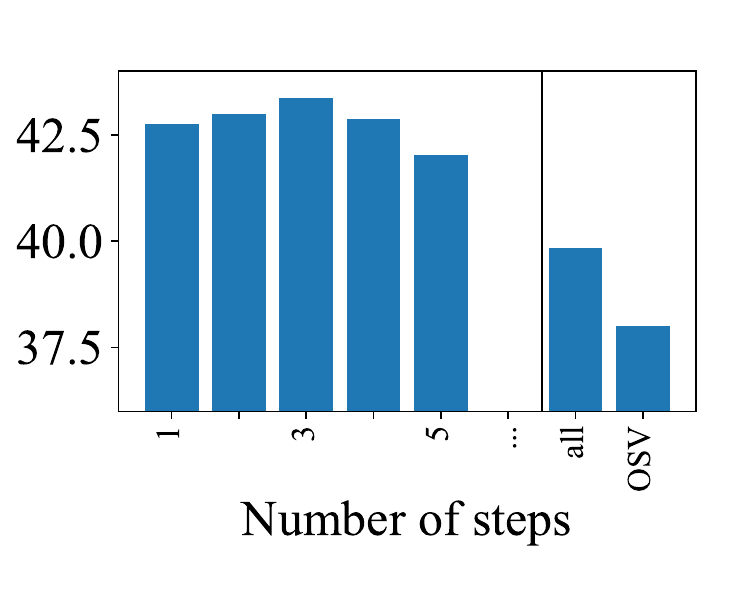}  
    }
    \subfigure[MBPP, \smallmodel, \sumlogit]{
        \includegraphics[width=.3\linewidth]{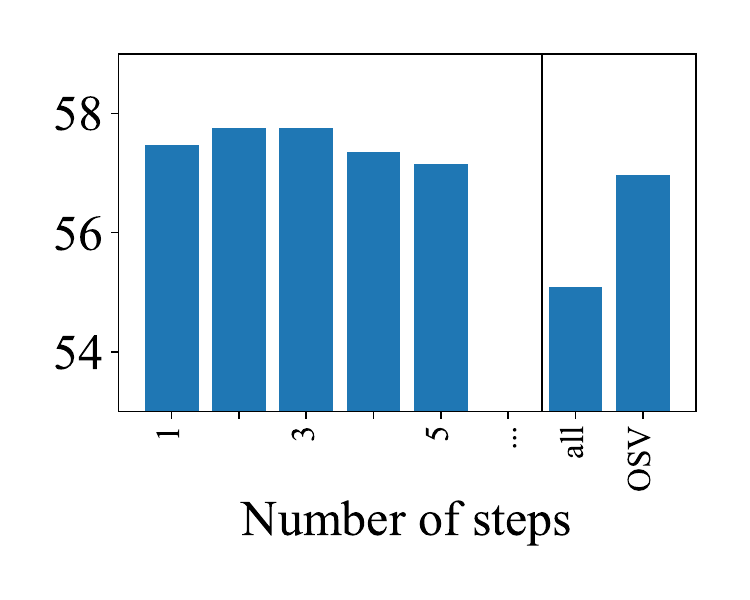}  
    }

    \subfigure[MATH, \largemodel, \sumlogit]{
        \includegraphics[width=.3\linewidth]{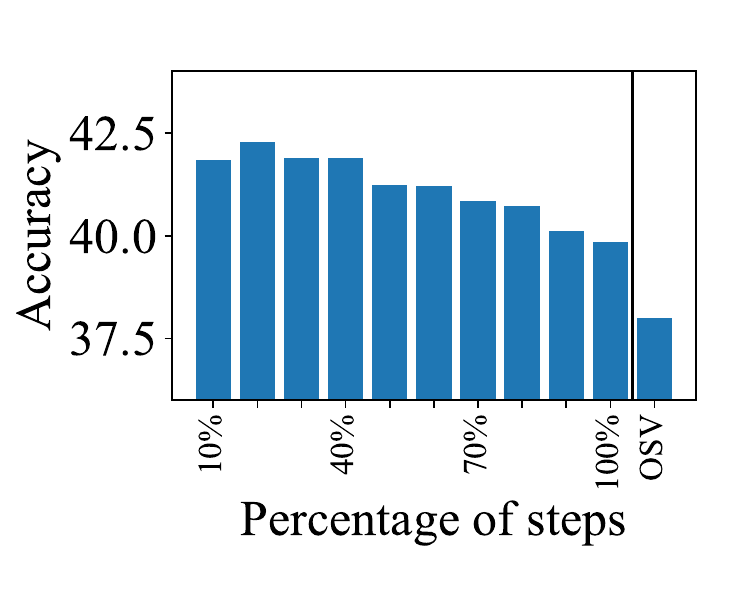}  
    }
    \subfigure[MATH, \largemodel, \maxv]{
        \includegraphics[width=.3\linewidth]{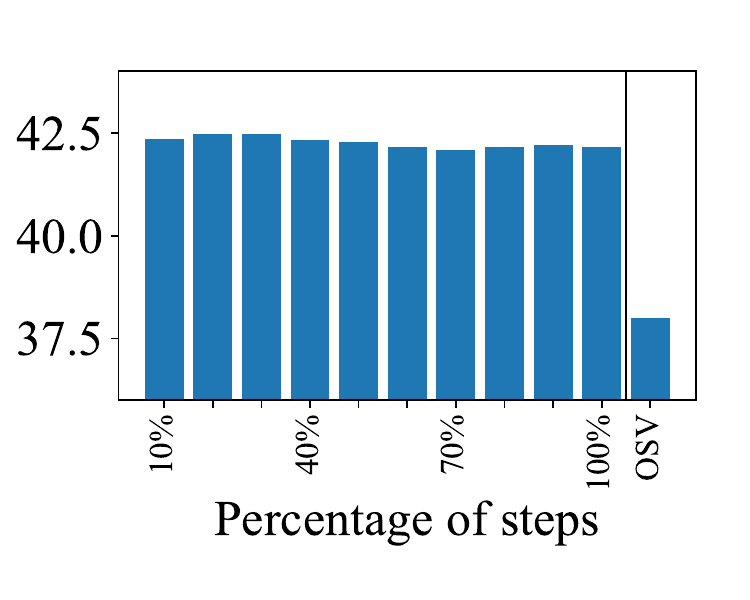}  
    }
    \subfigure[MATH, \largemodel, \sumodd]{
        \includegraphics[width=.3\linewidth]{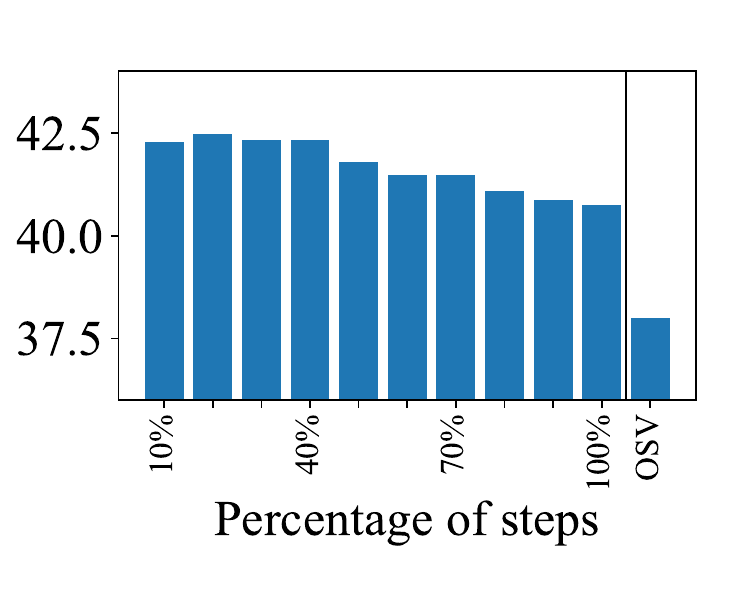}  
    }
    \caption{We plot the performance of using various aggregation functions with \processverifier while restricting it to predict only the last $k$ steps or the last $p$ percentage of steps.}
    \label{fig:length}
\end{figure*}

We first present the performance of the process verifier trained on \method data, using the default training objective on correctness scores directly, and the \maxv aggregation function on the three datasets. For this experiment, we varied the number of solutions to be verified by the verifier from 2 to 128, to clearly depict the trend of the compared verifier's performance. 
The results are shown in Fig.~\ref{fig:main}.
The plots convey several pieces of information. 
\begin{itemize}[nosep,leftmargin=*]
\item It is evident that using any verifier improves significantly upon no verification, matching with the initial assumption that verification plays a vital role in multi-step problem solving.
\item In all experiments, the verifier trained on \method using the max aggregation function showed stronger results than output verification. On GSM8K, the process verification is better than self-consistency. On MATH, the performance lacks a bit. We note that this may be because we are training a less competent verifier (\smallmodel) than the reasoner (\largemodel). \citet{mathshepherd} showed improvements upon self-consistency when the verifier and reasoner are of the same sizes. 
\item The high performance of \maxv may be unexpected, as \maxv seemly would be biased towards the first few correct steps of an incorrect solution. In later analysis, we will show that (1) \maxv favors high scores, similar to some other aggregation functions that perform well, that is preferred on \method data; (2) Due to higher noise in the earlier steps in \method data, the prediction scores of the earlier steps is of lower value (i.e., model confidence is lower), thus showing less affect to \maxv.
\item In all experiments, the \sumlogprob (product of probabilities) and \minv aggregation function are much worse than \maxv or even using \outputverifier, nevertheless still providing benefits over not using a verifier. 
\item For several verifiers, we observe that the performance of the verifier is on a decreasing trend when the number of generations is high. This is particularly interesting since the larger the generations, the closer the performance should approximate the true verification performance. This would indicate that the while the verifier might identify some correct solutions with high scores, it also incorrectly predicts some fewer incorrect solutions with even higher scores, a sign of improper generalization.
\end{itemize}

From these results, we focus our analysis on two subjects: (1) the choice of aggregation functions, and (2) the effect of noise on generalization of the \processverifier.

\subsection{Aggregation Functions}


\begin{table}[th]
    \small
    \centering
    \caption{We test on GSM8K the effect of different training objectives and aggregation functions (Hard $\&$ \minv , the combination used in \citet{mathshepherd}, and Soft $\&$ \maxv , the combination we suggest). All 3 base models are 7B in size, and MM denotes the MetaMath~\cite{metamath} fine-tuned version of them.}
    \label{tab:train_obj}
    \input{1-tables/train_obj}
\end{table}

\begin{table*}[th]
    \small
    \centering
    \caption{We train a verifier based on \smallmodel and data generated by \smallmodel and test its applicability to transfer to validate solutions generated by two different reasoners, \largemodel and \texttt{gpt-turbo-3.5}. A reference score of validating solutions generated by \smallmodel itself is also given. We evaluate this on two tasks, GSM8K and MBPP.}
    \label{tab:transfer}
    \input{1-tables/transfer}
\end{table*}

To start with the analysis, we consider ten aggregation functions (sum and means of log probabilities, probabilities, logits, and odds, and maximum and minimum value over all steps). 
We obtain their performance on the \method dataset and plot it with the performance of the verifier using the aggregation function on the test set (Fig.~\ref{fig:aggr}).
We first observe that in general, the two performances have positive correlations, indicating that it is possible to select an aggregation function on the \method dataset and use it during inference.
Second, we notice that both \minv and \sumlogprob have low performances not only during inference, but also in \method. 
This indicates that the poor performance of them likely is related to the construct of \method. 
Indeed, we realize that \sumlogprob does not have a high correlation with a correct solution, as it naturally penalizes long solutions. 
For \minv, the possibility for a set of solutions to be wrongly verified using \minv is when the solution with the largest minimum correctness over all steps turns out to be wrong. 
This is actually not an unlikely event to happen, particularly in consideration when the reasoner makes an erroneous continuation to an initially correct solution.
To clarify these better, we answer the following questions: 

\noindent\textbf{What are common in good aggregation functions for \method data?} We believe a rule of thumb of a good aggregation function is a function that \textbf{values high scores highly}. Consider two functions, one that values high scores (selects the solution with the highest high scores, e.g., \maxv) and one that values the low scores (selects the solutions with the highest low scores, e.g., \minv). The first function is wrong only when the highest score solution is incorrect, in a simple case where there is only one observed step score for each solution, the probability is $1 - s_{\text{max}},$ where $s_{\text{max}}$ is the score. Similarly, for \minv, the probability that the solution with the highest minimum score is wrong is $1 - s_{\text{min}}.$ Since $s_{\text{max}} \ge s_{\text{min}}, $ the first function shall be preferred. This is in line with the observation from Fig.~\ref{fig:aggr} that aggregations of odds and logits are usually better than that of probabilities and log probabilities.

\noindent\textbf{Why did \sumlogprob and \minv work well in \citet{stepbystep} and \citet{mathshepherd}?}  
In \citet{stepbystep}, the dataset is constructed by human identifying all (earliest) incorrect steps, which corresponds to a prediction of 0 for the verifier (i.e., following the analogy in the previous discussion, this indicates that $s_{\text{max}} = s_{\text{min}} = 1.0$). The \minv function would be correct on every instance in the training dataset, and if the verifier generalizes well, resembles human identification of mistakes on the test dataset. For \citet{mathshepherd}, we note a difference during \method data construction as their training objective is to predict the binary value of the correctness score, we discuss this more in Sec~\ref{sec:train_obj}.

The aggregation function analysis would indicate that a good \method dataset score indicates a good aggregation function.
This is not completely correct, since, a contradictory result is that the final step score, which is used to train the output supervised verifier, achieves 100\% accuracy on the training dataset, while not as good as the process supervised verifier on the test set. 
This suggests that the output supervised verifier might encounter some generalization issues from the data, and \method data can help relieve them. 

\subsection{Different Length Aggregations}

To understand the generalization issue, we illustrate the result of applying an aggregation function to only the last $k$ steps or last $p$ percentage steps of the solutions in Fig.~\ref{fig:length}. 
In the upper three plots, we show the performance of an aggregation function \sumlogit on the three datasets with $1 \leq k \leq 5$.
In the lower three plots, we show the performance of three aggregation functions on MATH with $10 \leq p \leq 100$. We only conducted this analysis on the MATH dataset, as it have solutions long enough such that looking at a percentage number of steps is sensible.

\begin{itemize}[leftmargin=*,nosep]
    \item For all experiments, the performance increases with a few more steps considered from the end. This indicates that the \processverifier predictions on the last steps brings in increasing value, suggesting that process scores indeed are beneficial.
    \item For most experiments, the performance starts to drop after including some early steps. This suggests that the quality of the predictions for the first steps are poor. We believe this is because \method has a poorer estimation of the first steps than the last steps, since intuitively it is hard to predict the correctness of a very early solution, causing burden for the verifier to learn. 
    \item For \maxv, the performance does not change significantly across including more earlier steps. We examined the predicted scores and find the usually earlier steps are smaller in value, causing it to contribute little to \maxv. This is another evidence that \processverifier trained on \method data might suffer from noise in the earlier steps.
    \item In all experiments, using the last-step process verifier predicted value is more beneficial than output supervision alone. Recall that this is not because of the problem of data quantity, as we upscaled the data to train the output verifier. We suggest that this is because the process supervision data is of more diverse context, thus helping the model in generalization. 
\end{itemize}

\subsection{Different Training Objectives}\label{sec:train_obj}
The main difference in the method of ours and \citet{mathshepherd} is the training objective of the verifier, where we train the verifier to directly predict the estimated accuracies (Soft Objective), and they train the verifier to predict a binarized value (non-zeroness) of the accuracy (Hard Objective). 
In our previous analysis, we noted that since the reasoner is imperfect, \method would provide underestimated accuracies of the intermediate steps, which is harmful to aggregation functions that focus on low values (e.g., \minv). In contrast, the non-zeroness of the accuracy would cause an overestimation of the accuracy, which, by the same argument, would be harmful to aggregation functions that focus on high values (e.g., \maxv). To verify this, we conduct the experiments using the same language model as \citet{mathshepherd} on the GSM8K dataset, using both training objectives and aggregation functions. 

The experiment setting is detailed in Appendix~\ref{app:settings2}. The results are in Table~\ref{tab:train_obj}. It is observed that, indeed, the max aggregation is better for the soft objective and the min aggregation is better for the hard objective. It also turns out that soft objective with the max aggregation consistently outperforms hard objective with min aggregation. We believe this to be a strong motivation for the use of the soft objective in \method.

\subsection{Transferring to a Different Reasoner}
Finally, we provide an auxiliary experiment to check whether the trained verifiers would transfer to different reasoning models. 
We apply the verifiers trained on reasoning data generated by a \smallmodel and use it to valid solutions generated by stronger reasoners (reasoners having higher \textit{No Verifier} accuracy). 
We find the \sumlogit aggregation function working well in this case. 
The result is shown in Tab.~\ref{tab:transfer}, which shows that the trained verifier transfers to different and stronger reasoners with a strong validation ability, indicating that the verifier is not learning something overly specific to the reasoner that generates the data.

%% file: 1-tables/train_obj.tex
\begin{tabular}{l|cccc}
\hline
Model         & Llemma & MM-Llemma  & MM-Mistral \\
\hline
Soft + \maxv  & 54.7 & 72.4 & 80.3 \\
Soft + \minv  & 51.2 & 70.1 & 77.8 \\
Hard + \maxv  & 50.2 & 68.9 & 78.1 \\
Hard + \minv  & 52.4 & 70.8 & 79.2 \\
\hline
\end{tabular}

%% file: 1-tables/transfer.tex
\begin{tabular}{l|cccc}
\hline
GSM8K \smallmodel          & No Verifier & Self Consistency & \outputverifier & \processverifier w/ \sumlogit \\
\hline
$\to$ \smallmodel         & 61.6 & 78.7 & 89.5 & 90.5 \\
$\to$ \largemodel         & 80.7 & 89.4 & 92.1 & 92.6 \\
$\to$ \texttt{gpt-turbo-3.5}  & 72.5 & 86.2 & 88.0 & 89.1 \\
\hline
MBPP \smallmodel           & No Verifier & \outputverifier & \processverifier w/ \sumlogit & \processverifier w/ \sumlogit (last 3 steps) \\
\hline
$\to$ \smallmodel         & 41.7 & 56.8 & 54.2 & 57.8 \\
$\to$ \largemodel         & 42.4 & 56.6 & 55.0 & 57.4 \\
$\to$ \texttt{gpt-turbo-3.5}  & 66.2 & 67.6 & 67.6 & 68.2 \\
\hline
\end{tabular}

%% file: 0-contents/50-conclusion.tex
\section{Conclusion}
In this work, we introduce \method to automatically annotate intermediate solutions for multi-step problem solving. 
Such data can be used to train a process supervised verifier that validates solutions generated by a reasoner. 
On two math datasets and one coding dataset, we demonstrated that \method improves the ability of picking the correct solution over an otherwise trained output supervised verifier.
We conduct analysis on the aggregation function used to pick the solution and suggest that compared to verifiers trained on human-annotated process supervision, \method data trained verifiers prefer different aggregation functions. 
We also showed that such verifiers do not overly emphasize on the mistakes of the reasoner that produced the data, and can be transferred to different reasoners. 
Future work could explore creating a scalable way to obtain \method data for each token in solutions to train a more competent verifier and use it to tune the reasoner via reinforcement learning.

\section{Limitation}

\subsection{Underperformance on the MATH dataset}
In our work, we did not manage to conduct all experiments using the same, large model. Especially for the MATH dataset, we had to train a smaller verifier to compensate of the long sequence length and data size. This probably led to us finding a lower performance of process and output verifier than the straightforward self-consistency. We believe in general that this is not true, as \citet{stepbystep} and \citet{mathshepherd} both showed that process/output verifier should output self-consistency on the MATH dataset.

\subsection{Efficiency}
\method, while automatic, requires a non-trivial amount of computation effort in generating the dataset to train the verifier. We did not attempt to reduce the computational effort, as we'd like to show the most direct comparison with no verifiers and output supervised verifiers. We do believe it is very possible to reduce the computation costs, for example, by avoiding creating data on every intermediate solutions, and we suggest future work to explore this direction.



%% file: 0-contents/appendix.tex
\section{Datasets}\label{app:datasets}
\begin{itemize}[leftmargin=*,nosep]
    \item \textbf{GSM8K}~\cite{gsm8k} is a dataset of grade school math problems. The solution is given in a one-line-per-step format with an exact numerical answer in the last line in the format of $\#\#\#\# \{\text{answer}\}.$ To enforce the reasoner following this format, we use the first 2000 instances in its training set to fine-tune the reasoner model to follow such a format. The solutions to fine-tune are from the training set. We use the coming $5000$ data to train the verifier and evaluate the verifier on solutions generated by the reasoner on the test set.
    \item \textbf{MATH}~\cite{math} is also a math word problems dataset. It consists of math problems of high school math competitions. The solutions are given in a format that mixes latex code and natural language. A dedicated solution checker was developed~\cite{math, stepbystep}. While the dataset itself does not resemble steps into different lines, we prompted GPT-4 to break down the reference solutions into one step per line, and fine-tuned the reasoner on the line separated dataset to make it follow the format. We use the test split suggested in \citet{stepbystep}.
    \item \textbf{MBPP} is an entry-level Python programming dataset. The questions are coding challenges along with a test case that defines the function format and the solutions are Python code that is expected to solve several hidden test cases. We treat each individual line in the generated code as a step. For languages like Python,  this resembles one statement per step. Due to the small dataset size, we can not afford to fine-tune the reasoner model, and decide to use 3 prompts in the validation split as in context examples to make sure the model generates code in the expected format.
\end{itemize}

\subsection{Settings}\label{app:settings}
Throughout the paper, we choose to use a temperature value of $r_g = r_{mc} = 0.7$ for both constructing \method and generating solutions on the test set. The number of generations for constructing \method $n_g$ is set to 32 for GSM8K and MBPP, and 8 for MATH. The number of completions $n_{mc}$ is also 32 for GSM8K and MBPP, and 8 for MATH.
For GSM8K, we experiment with both \smallmodel and \largemodel in the reasoner-verifier framework.
For MATH, due to compute constraints, the reasoner we use is the \largemodel, and the verifier trained is the \smallmodel.
For MBPP, we find marginal differences in performance between using \smallmodel and \largemodel as the reasoner, therefore we experiment only with the \smallmodel. 
During generation, all models are 8-bit quantized, and during training, we use a bfloat16 precision.
Since \method contains an annotation for each intermediate step in the solution, it is naturally the number of steps times larger than output supervision. Therefore, we additionally generate more data to train the \outputverifier.
For training, we follow standard reward model training recipes, with an exception on the training epochs. Similar to \citet{stepbystep}, we also find it better to train the \outputverifier for 1 epoch and the \processverifier for 2 epochs. For the \outputverifier on MATH data, we find that training with a small 0.2 epochs (essentially training on less data) is better than training longer. 
For all experiments, we report the results of the average of 5 independently trained verifiers with different random seeds.

\subsection{Settings of the Objective Experiment}\label{app:settings2}
To reduce the cost, when conducting the experiment to compare the two training objectives, we scaled down the experiment. On GSM8K, we used 2000 data points for verification training data generation, and $n_g = n_{mc} = 8$.